\pdfoutput=1
\documentclass{article}
\usepackage{spconf,amsmath,graphicx}
\usepackage{tikz,lipsum}

\usepackage{subcaption}
\usepackage{color}

\newcommand\blfootnote[1]{
  \begingroup
  \renewcommand\thefootnote{}\footnote{#1}
  \addtocounter{footnote}{-1}
  \endgroup
}

\newcommand\copyrighttext{
\footnotesize Copyright \textcopyright 2021 IEEE. Published in the IEEE 2021 International Conference on Image Processing (IEEE ICIP 2021), scheduled for 19-22 September 2021 in Anchorage, Alaska, United States. Personal use of this material is permitted. However, permission to reprint/republish this material for advertising or promotional purposes or for creating new collective works for resale or redistribution to servers or lists, or to reuse any copyrighted component of this work in other works, must be obtained from the IEEE. Contact: Manager, Copyrights and Permissions / IEEE Service Center / 445 Hoes Lane / P.O. Box 1331 / Piscataway, NJ 08855-1331, USA. Telephone: + Intl. 908-562-3966.
}
\newcommand\IEEEnotice{
\begin{tikzpicture}[remember picture,overlay]
\node[anchor=south,yshift=10pt] at (current page.south) {\fbox{\parbox{\dimexpr\textwidth-\fboxsep-\fboxrule\relax}{\copyrighttext}}};
\end{tikzpicture}
}

\title{FAST HYBRID IMAGE RETARGETING}
\name{
  Daniel Valdez-Balderas,
  Oleg Muraveynyk,
  Timothy Smith
}
\address{Samsung R\&D Institute, Staines-upon-Thames, TW18 4QE, U.~K.}

\begin{document}
\maketitle
\begin{abstract}
Image retargeting changes the aspect ratio of images while aiming to preserve content and minimise noticeable distortion. Fast and high-quality methods are particularly relevant at present, due to the large variety of image and display aspect ratios.
We propose a retargeting method that quantifies and limits warping distortions with the use of content-aware cropping. The pipeline of the proposed approach consists of the following steps.
First, an importance map of a source image is generated using deep semantic segmentation and saliency detection models.
Then, a preliminary warping mesh is computed using
axis aligned deformations \cite{AxisAligned:2012}, enhanced with the use of a distortion measure to ensure low warping deformations.
Finally, the retargeted image is  produced using a content-aware cropping algorithm.
In order to evaluate our method, we perform a user study based on the RetargetMe benchmark. Experimental analyses show that our method outperforms recent approaches, while running in a fraction of their execution time.

\end{abstract}


\blfootnote{The authors wish to thank Mete Ozay for his advice and support.}
\IEEEnotice
\section{Introduction}
\label{sec:intro}

The use of visual media on mobile devices has seen dramatic increases over the last two decades. At the same time, the variety of displays continues to evolve and now includes dynamically changing factors, such as those in foldable phones. This multiplicity of media and display sizes makes retargeting a highly relevant technology. Retargeting is a process by which the aspect ratio of an image is changed, while aiming to keep important content in the image and to minimise noticeable distortion \cite{VisualMediaRetargeting:2009}. In the era of mobile computing, fast retargeting methods, which allow real-time, interactive usage on mobile devices, are of particular importance
\cite{RetargetingForMobileDevices2018}.

\begin{figure}[tpb]
      \centering
      \centerline{\includegraphics[width=8.5cm]{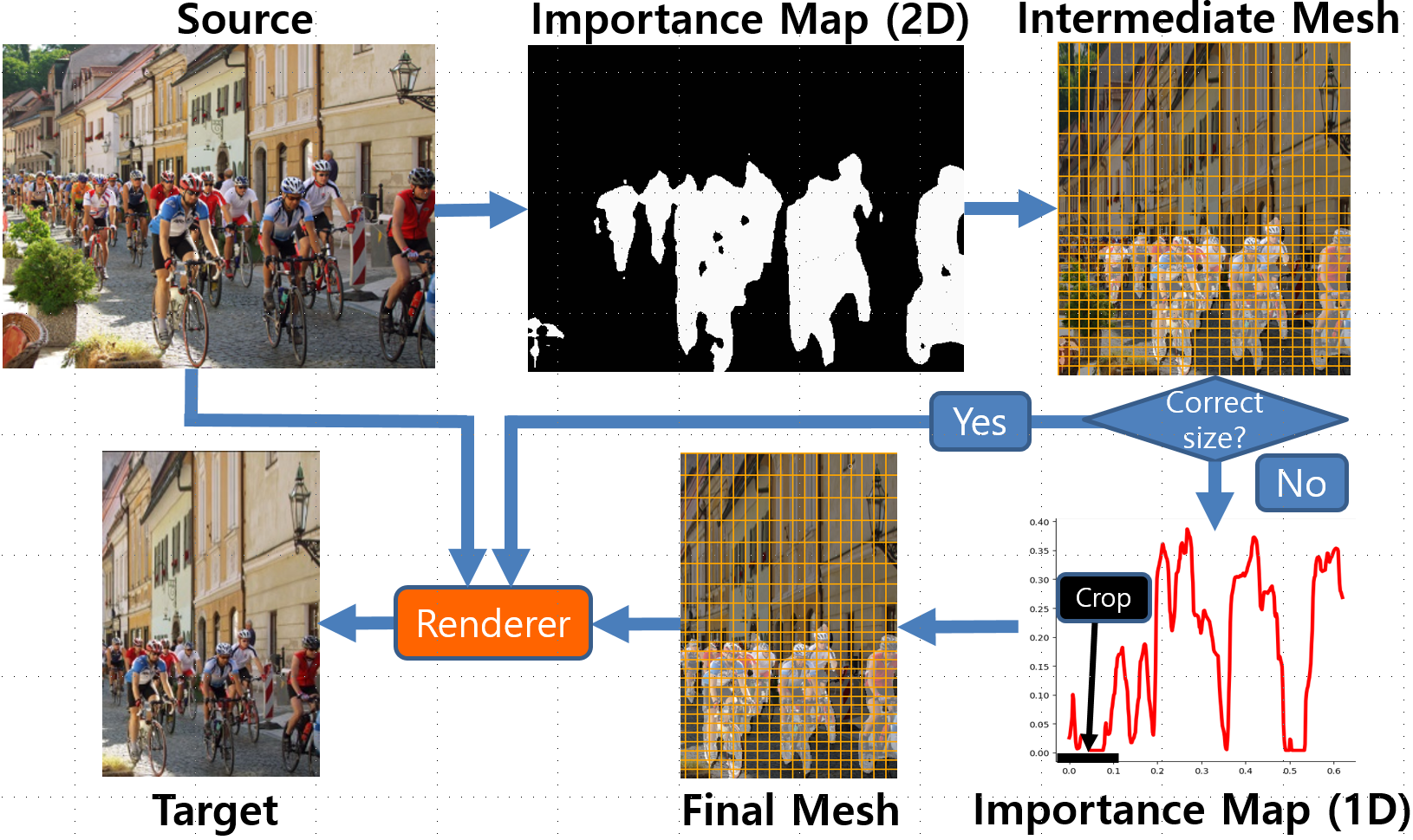}}
    \caption{An overview of the proposed retargeting method.}
    \label{fig:pipeline_full}
\vspace{-0.5cm}
\end{figure}

Combining the capabilities of existing Deep Neural Network (DNN) architectures with the efficiency of a warping algorithm \cite{AxisAligned:2012},
this work presents a novel method for image retargeting.
Our implementation has been tested both on desktop and on mobile devices, and a user study based on the RetargetMe benchmark
\cite{RetargetMe:2010}
shows that our method outperforms recent approaches, while executing in a fraction of the time
\cite{selfplay2020, CycleIR2019, resizingByReconstruction2019, weaklyAndSelfSupervised2017, beltrami, ahmadi2019contextaware}.

The contributions of this work are as follows.
First, we demonstrate that DNNs can be successfully used to produce importance maps required by a retargeting approach based on warping \cite{AxisAligned:2012}.
Second, we propose a novel, simple, and effective way to quantify distortions, and to limit them with the use of content aware-cropping.
Third, we demonstrate via a user study that our method
produces state-of-the-art results.
Fourth, we demonstrate the feasibility of retargeting on mobile devices, which can be done interactively in real-time once a pre-processing step has been performed.

\begin{figure}[t]
    \centering
    \begin{subfigure}[b]{0.95\linewidth}
        \centering
        \includegraphics[width=6.5cm]{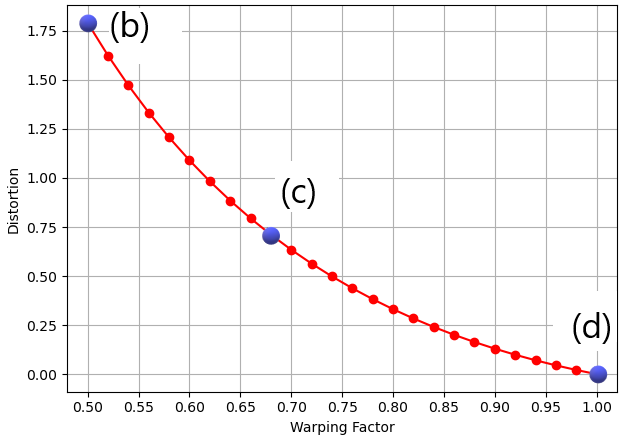}
        \caption{Change of distortion with respect to the warping factor.}
    \end{subfigure}

    \begin{subfigure}[b]{.3\linewidth}
        \centering
        \includegraphics[width=2.50cm]{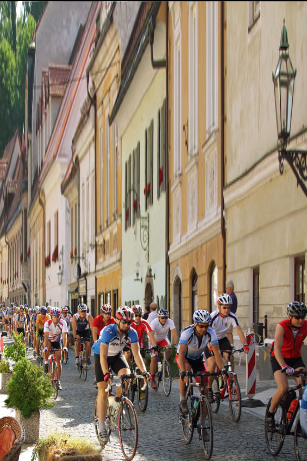}
        \caption{Warp only.}
    \end{subfigure}
    \begin{subfigure}[b]{.3\linewidth}
        \centering
        \includegraphics[width=2.50cm]{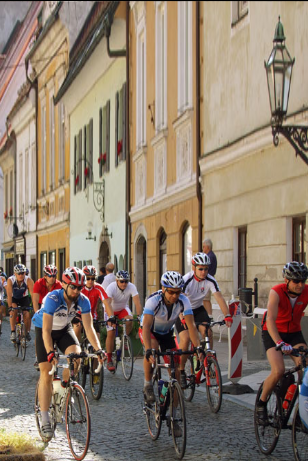}
        \caption{Hybrid.}
    \end{subfigure}
    \begin{subfigure}[b]{0.3\linewidth}
        \centering
        \includegraphics[width=2.50cm]{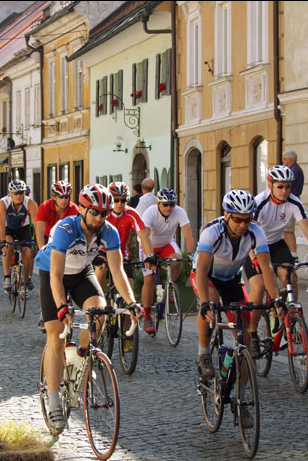}
        \caption{Crop only.}
    \end{subfigure}
 \caption{A measure is proposed to quantify and limit warping distortions. (a) As an image is warped to reduce its width,
 background distortions $D$ increase. (b) An image retargeted using only warping, where houses are deformed largely. (c) An image retargeted using the hybrid warp-crop approach, which reduces distortions in the houses as compared to warping-only, while preserving most important content. (c) An image retargeted by cropping only, where much of the original content has been lost, but has no distortions.
}
\label{fig:distortion}
\vspace{-0.5cm}
\end{figure}
\section{Preliminaries}
\label{sec:preliminaries}
Image retargeting approaches have been categorised as either discrete or continuous
\cite{VisualMediaRetargeting:2009}.
Discrete approaches, like seam carving
\cite{AS07,Rubinstein08},
add or remove pixels from source images to achieve the target size. Continuous methods change image dimensions by finding a mapping (a warp) from source to target images
\cite{ASystemForVideoRet, EnergyBasedDeformation, ByQuadraticProgramming}.
More recently, a variety of retargeting algorithms
\cite{
  CycleIR2019,
  resizingByReconstruction2019,
  weaklyAndSelfSupervised2017,
  beltrami,
  ahmadi2019contextaware,
  RetargetingForMobileDevices2018,
  continuousSeamCarving2020,
  composingSemanticCollage2018,
  deepIR2019}
and on reinforcement learning
\cite{selfplay2020}
have been proposed.
In this section, we describe prior art which is of particular relevance to our method.

One of the building blocks of our approach is Axis Aligned Deformations (AAD)~\cite{AxisAligned:2012}. Given an importance map,
AAD finds the optimal warping by optimizing an energy function of mesh coordinates, and uses the map to penalise distortions of important regions.
In AAD, the mesh is defined by a set of vertical and horizontal lines, illustrated by the Intermediate Mesh in Fig.~\ref{fig:pipeline_full}.
AAD produced
high quality results in a study~\cite{AxisAligned:2012} that compared \textit{non-fully automated} methods, i.~e. methods that use hand-crafted importance maps or parameters hand-tuned on a per-image basis. In this work, by contrast, AAD will be used to build a \textit{fully automated} retargeting system which does not require any human input.

In \cite{Rubinstein09-multiop}, three operators, namely cropping, seam carving, and uniform scaling, are integrated. Their algorithm finds an optimal path in the resizing space (i.~e., the space spanned by the three operators), given a global objective function that measures the similarity between source and retargeted images.
Our proposed approach differs from that work both in terms of the operators used, as well as on the criteria which governs the usage of the operators. Here the operators are AAD-based warping and content-aware cropping, and they are employed considering a novel distortion measure based on importance maps, rather than using a similarity function.

\section{Methodology}
\label{sec:methodolgoy}

An overview of our method is illustrated in Fig.~\ref{fig:pipeline_full}. It consists of five stages: (1) importance map generation, (2)  intermediate warping mesh generation, (3) determination of cropping regions, if any, (4) generation of final warping-cropping mesh, and (5) final rendering.
An importance map is produced using semantic segmentation and saliency detection. This map is fed to AAD to produce an intermediate warping mesh. If this mesh matches the target size, then it is used to produce the target and the algorithm stops.
Otherwise,
content-aware cropping is performed, using a one dimensional version of the  map.

\subsection{Importance Map Generation}
\label{ssec:importance_map}

We generate an importance map using deep semantic segmentation \cite{deeplabv3plus2018} and saliency detection models \cite{SaliencyContextCodec2020}. First, a source image is fed to a segmentation model. Then, a test is applied to determine significance of detections provided by the model: if the percent of pixels detected as non-background objects is above a certain threshold, then the algorithm assigns an importance score of 1 to pixels with detections, and 0 otherwise.  However, if the percent of detections is less than the threshold, then the source image is put through a saliency detector, and the saliency score is used as the importance score.
An example of an importance map is shown in Fig.~\ref{fig:pipeline_full}.

\begin{figure}[tpb]
  \noindent
  \begin{minipage}[b]{0.270\linewidth}
    \centering Source
  \end{minipage}
  \begin{minipage}[b]{.135\linewidth}
    \centering SCL
  \end{minipage}
  \begin{minipage}[b]{0.135\linewidth}
    \centering BEL
  \end{minipage}
  \begin{minipage}[b]{0.135\linewidth}
    \centering CycleIR
  \end{minipage}
  \begin{minipage}[b]{0.135\linewidth}
    \centering AAD
  \end{minipage}
  \begin{minipage}[b]{0.135\linewidth}
    \centering Ours
  \end{minipage}
  \begin{minipage}[b]{0.270\linewidth}
    \centering
    \centerline{\includegraphics[width=2.4cm]{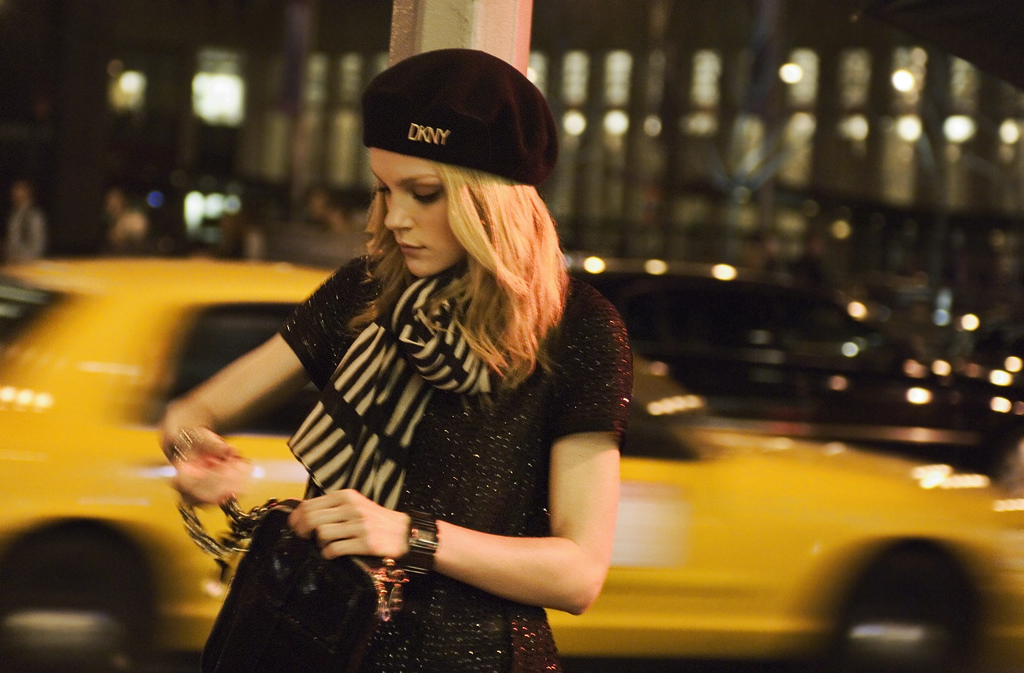}}
  \end{minipage}
  \begin{minipage}[b]{0.135\linewidth}
    \centering
    \centerline{\includegraphics[width=1.2cm]{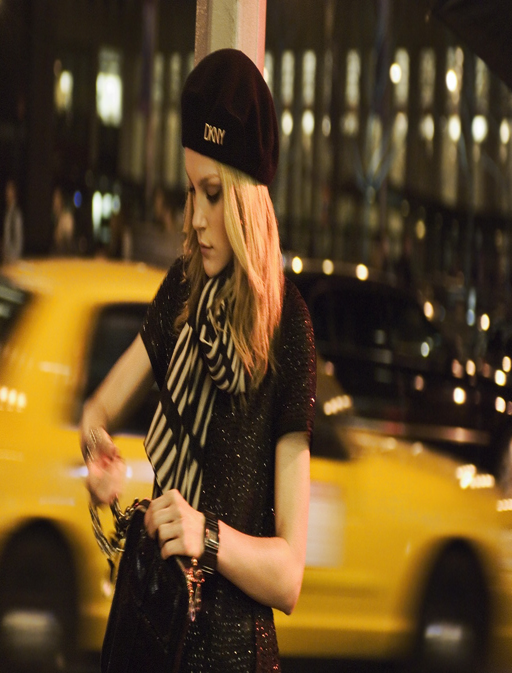}}
  \end{minipage}
  \begin{minipage}[b]{0.135\linewidth}
    \centering
    \centerline{\includegraphics[width=1.2cm]{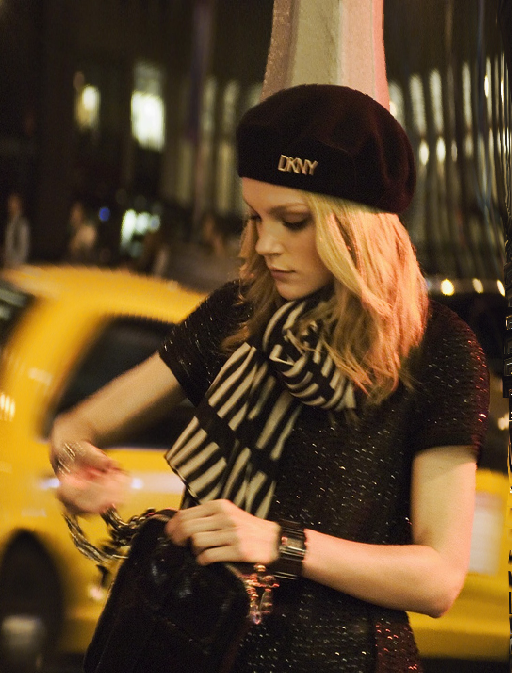}}
  \end{minipage}
  \begin{minipage}[b]{0.135\linewidth}
    \centering
    \centerline{\includegraphics[width=1.2cm]{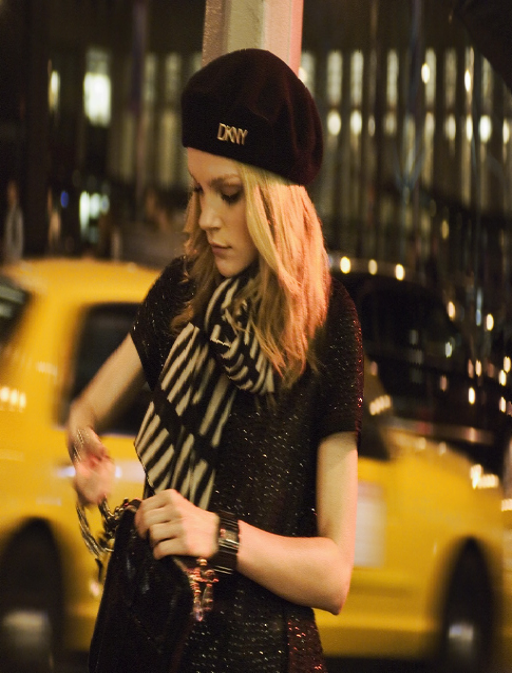}}
  \end{minipage}
  \begin{minipage}[b]{0.135\linewidth}
    \centering
    \centerline{\includegraphics[width=1.2cm]{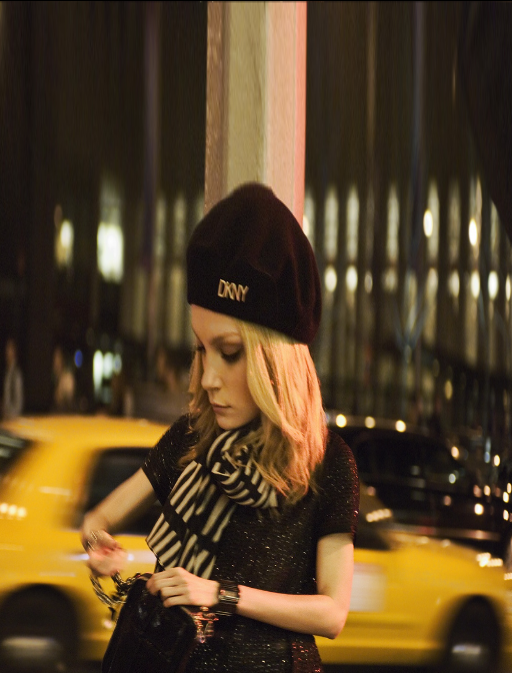}}
  \end{minipage}
  \begin{minipage}[b]{0.135\linewidth}
    \centering
    \centerline{\includegraphics[width=1.2cm]{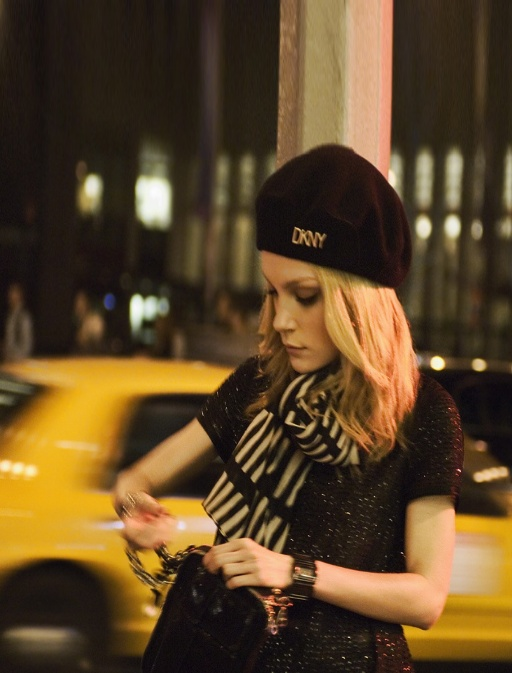}}
  \end{minipage}
  \begin{minipage}[b]{0.270\linewidth}
    \centering
    \centerline{\includegraphics[width=2.4cm]{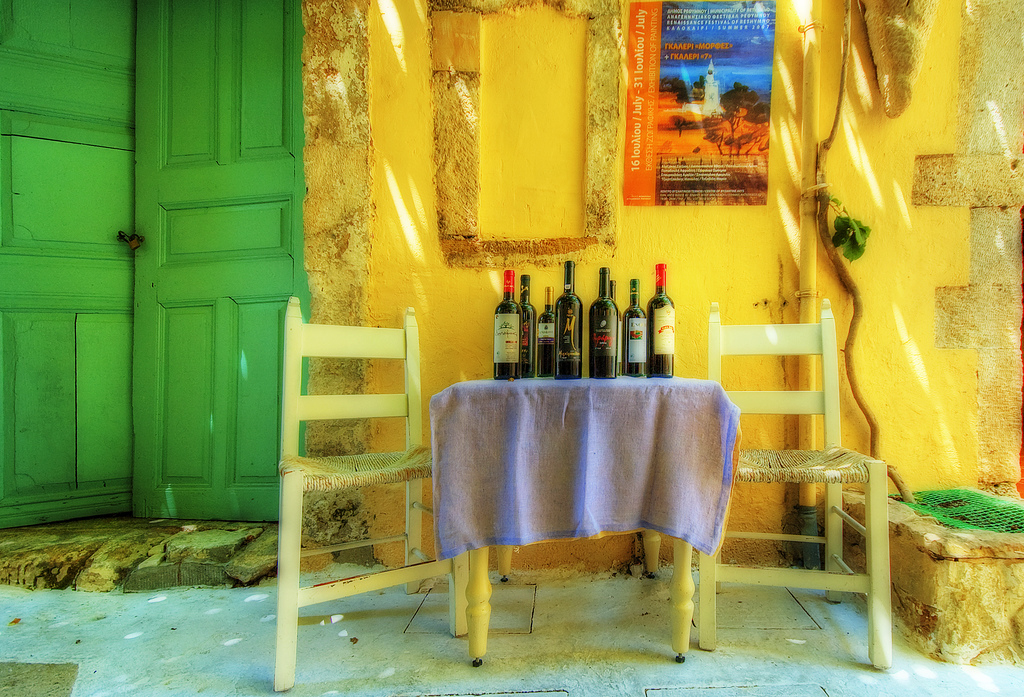}}
  \end{minipage}
  \begin{minipage}[b]{0.135\linewidth}
    \centering
    \centerline{\includegraphics[width=1.2cm]{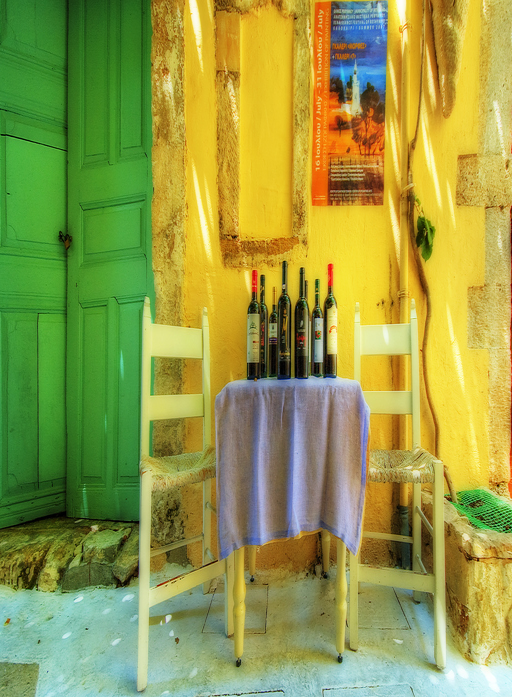}}
  \end{minipage}
  \begin{minipage}[b]{0.135\linewidth}
    \centering
    \centerline{\includegraphics[width=1.2cm]{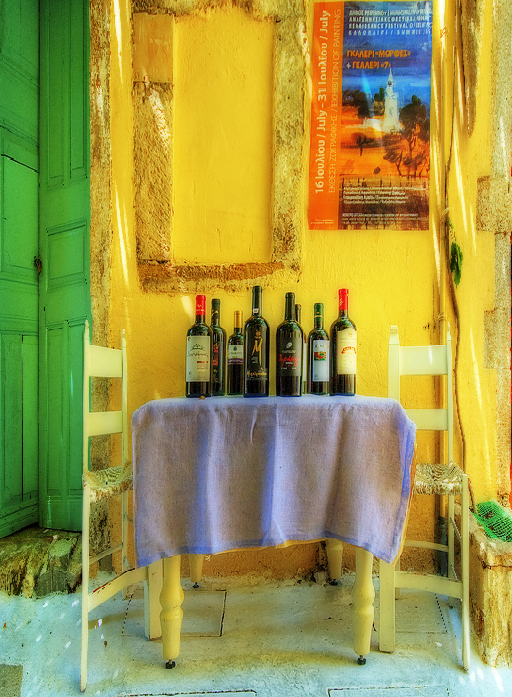}}
  \end{minipage}
  \begin{minipage}[b]{0.135\linewidth}
    \centering
    \centerline{\includegraphics[width=1.2cm]{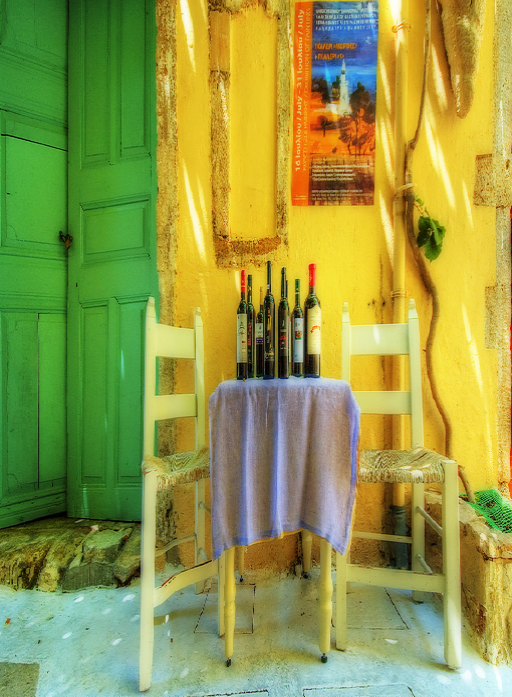}}
  \end{minipage}
  \begin{minipage}[b]{0.135\linewidth}
    \centering
    \centerline{\includegraphics[width=1.2cm]{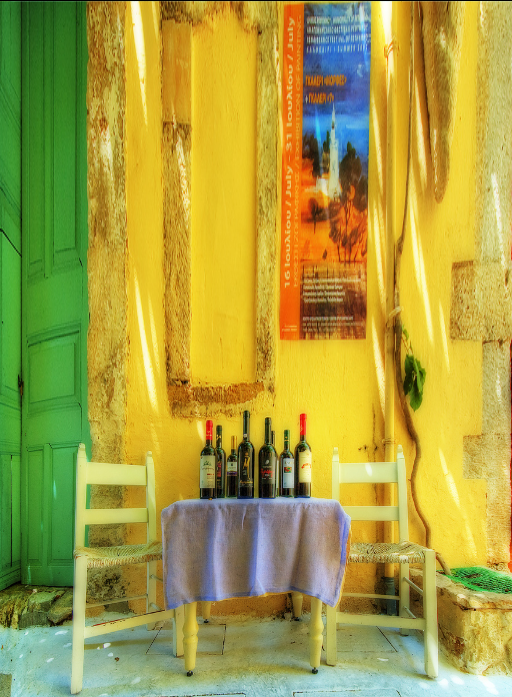}}
  \end{minipage}
  \begin{minipage}[b]{0.135\linewidth}
    \centering
    \centerline{\includegraphics[width=1.2cm]{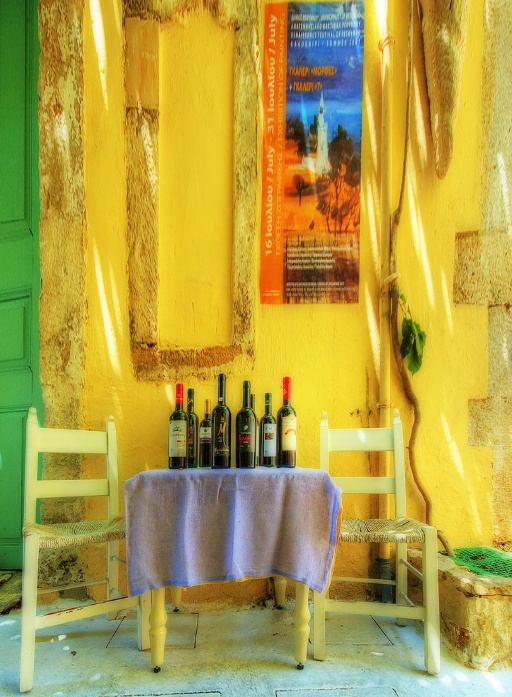}}
  \end{minipage}
  \begin{minipage}[b]{0.270\linewidth}
    \centering
    \centerline{\includegraphics[width=2.4cm]{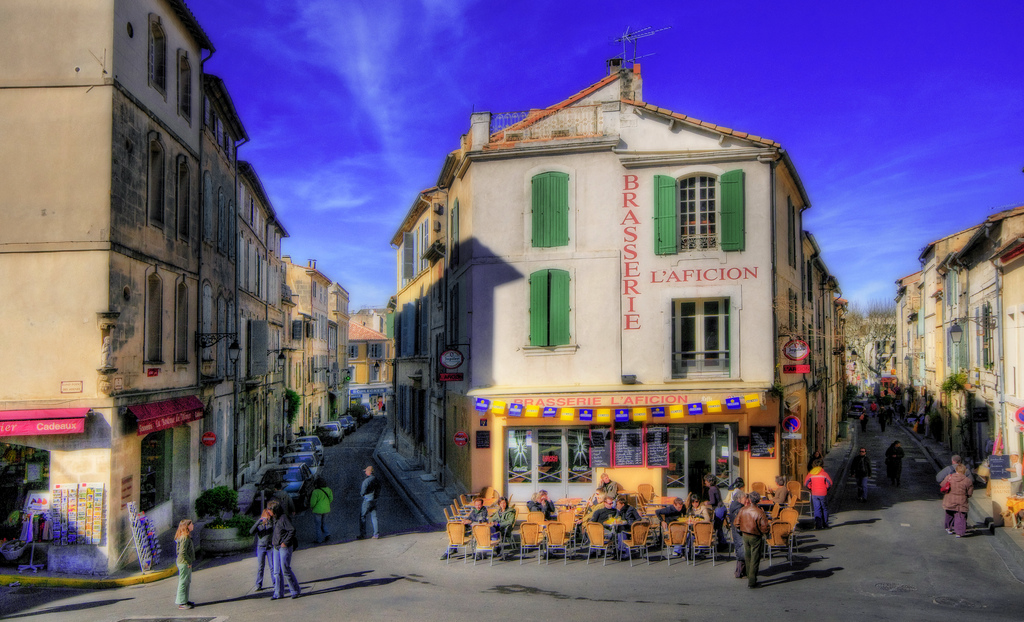}}
  \end{minipage}
  \begin{minipage}[b]{0.135\linewidth}
    \centering
    \centerline{\includegraphics[width=1.2cm]{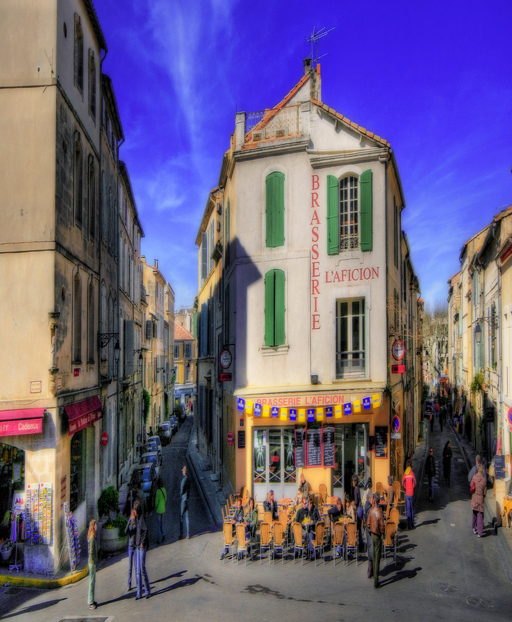}}
  \end{minipage}
  \begin{minipage}[b]{0.135\linewidth}
    \centering
    \centerline{\includegraphics[width=1.2cm]{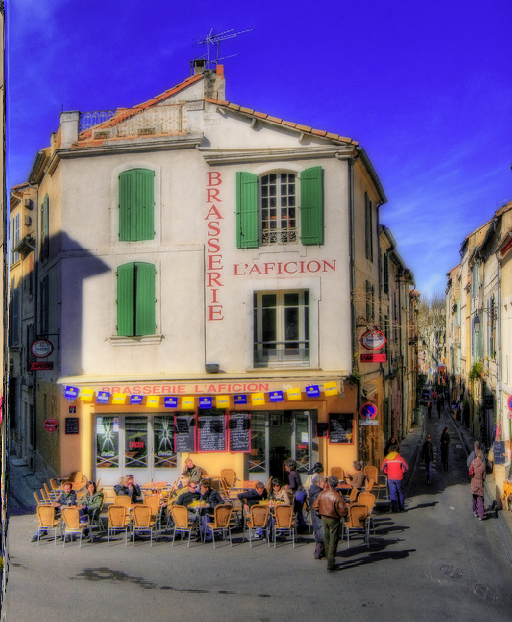}}
  \end{minipage}
  \begin{minipage}[b]{0.135\linewidth}
    \centering
    \centerline{\includegraphics[width=1.2cm]{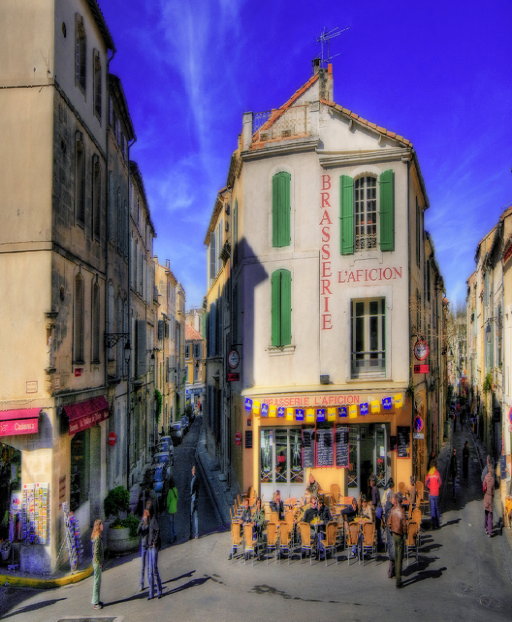}}
  \end{minipage}
  \begin{minipage}[b]{0.135\linewidth}
    \centering
    \centerline{\includegraphics[width=1.2cm]{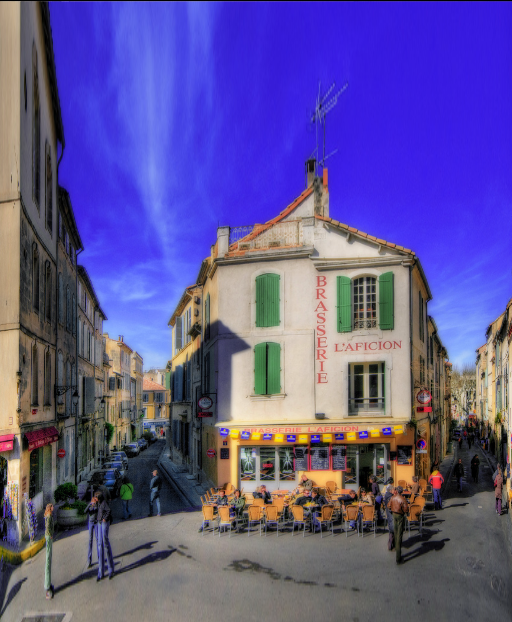}}
  \end{minipage}
  \begin{minipage}[b]{0.135\linewidth}
    \centering
    \centerline{\includegraphics[width=1.2cm]{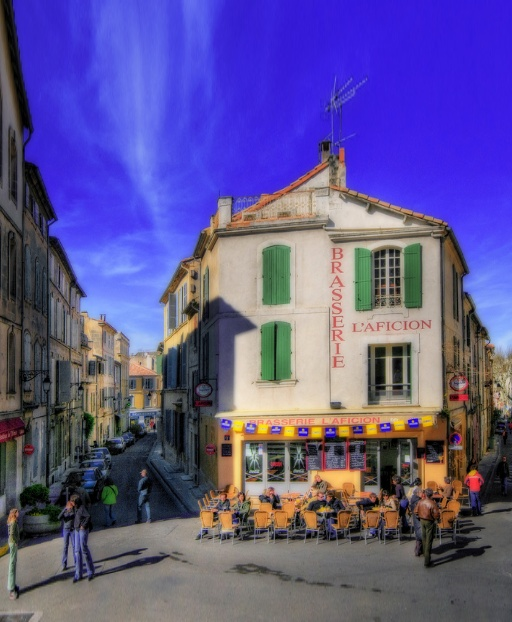}}
  \end{minipage}
  \begin{minipage}[b]{0.270\linewidth}
    \centering
    \centerline{\includegraphics[width=2.4cm]{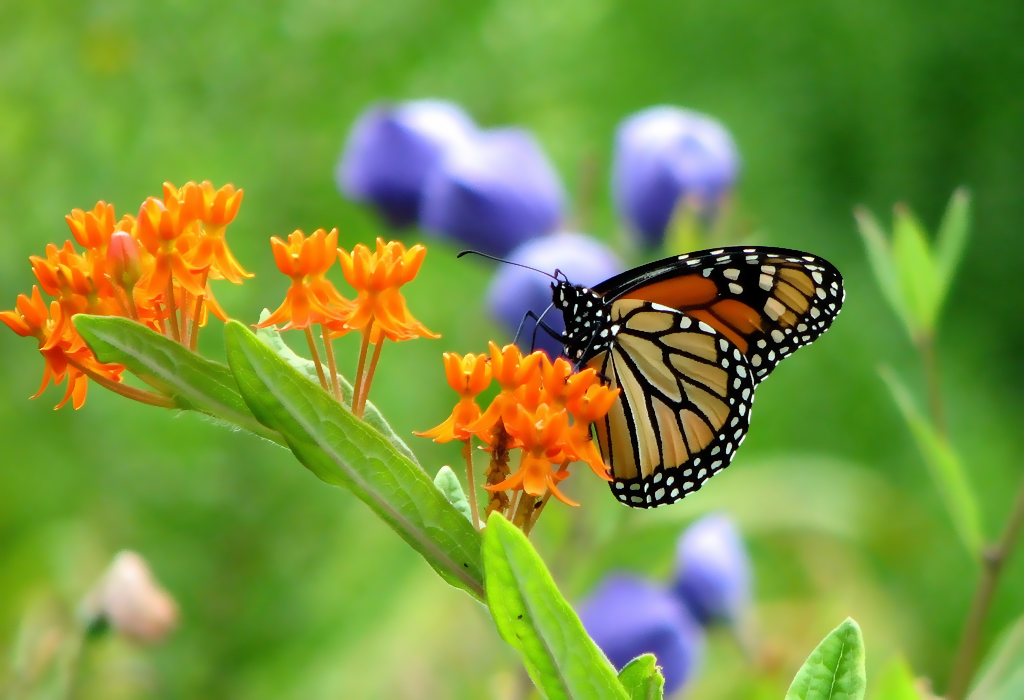}}
  \end{minipage}
  \begin{minipage}[b]{0.135\linewidth}
    \centering
    \centerline{\includegraphics[width=1.2cm]{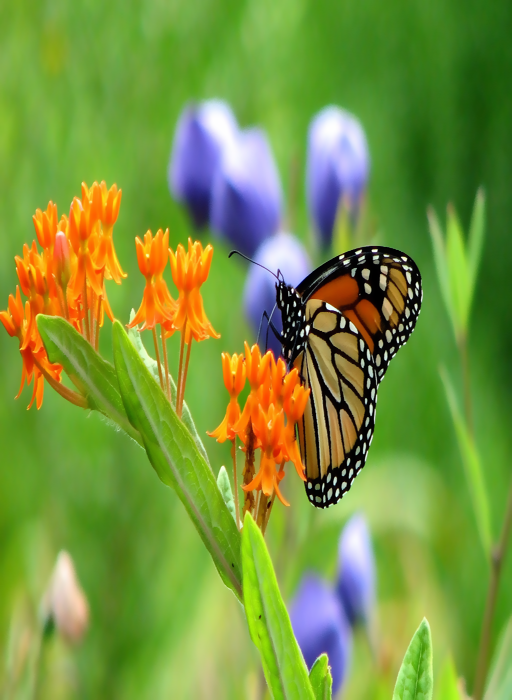}}
  \end{minipage}
  \begin{minipage}[b]{0.135\linewidth}
    \centering
    \centerline{\includegraphics[width=1.2cm]{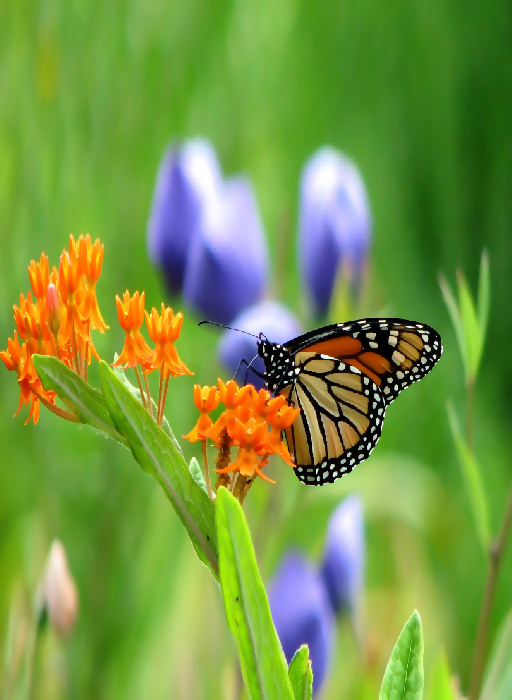}}
  \end{minipage}
  \begin{minipage}[b]{0.135\linewidth}
    \centering
    \centerline{\includegraphics[width=1.2cm]{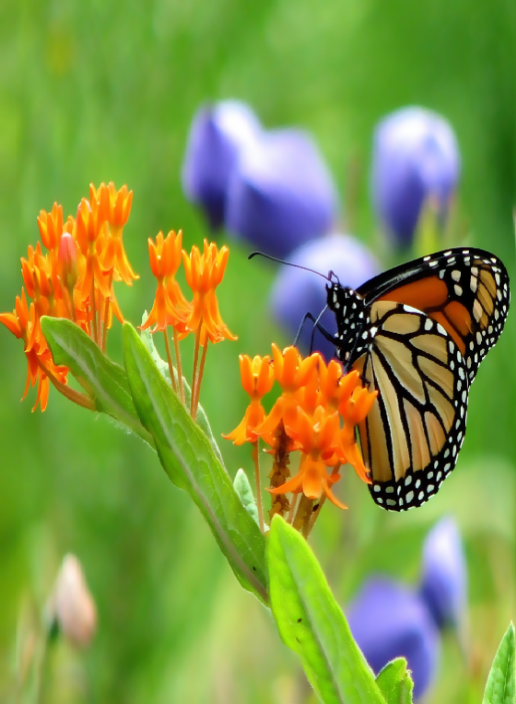}}
  \end{minipage}
  \begin{minipage}[b]{0.135\linewidth}
    \centering
    \centerline{\includegraphics[width=1.2cm]{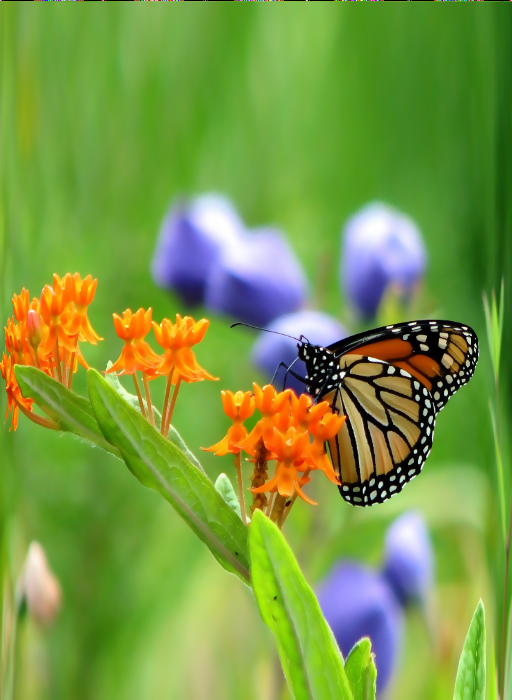}}
  \end{minipage}
  \begin{minipage}[b]{0.135\linewidth}
    \centering
    \centerline{\includegraphics[width=1.2cm]{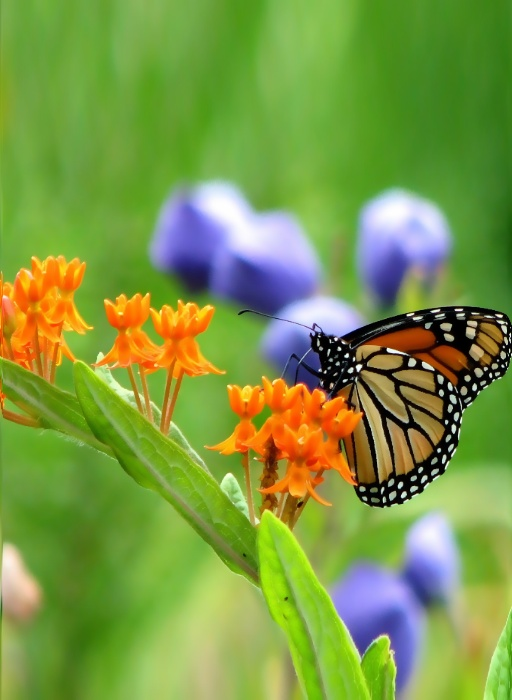}}
  \end{minipage}
  \caption{Source images, and their corresponding retargeted versions obtained using different methods. Our method (rightmost column) attains the best balance between keeping important content and a low level of distortion.}
  \label{fig:comparisons}
  \vspace{-0.5cm}
\end{figure}

\subsection{Intermediate Warping}
\label{ssec:intermediate_warping}

After an importance map is produced, an intermediate warping mesh is obtained via an  algorithm that finds the optimal warping that AAD \cite{AxisAligned:2012} can produce. The optimal warp is defined as the AAD warp that, when applied to the source image would produce an image that is as close to the target size as possible, but which does not introduce excessive distortions. To this end, a measure that quantifies distortions is needed.

\begin{figure}[tpb]
  \noindent
  \begin{minipage}[b]{0.594\linewidth}
    \centering Source
  \end{minipage}
  \begin{minipage}[b]{.195\linewidth}
    \centering AAD
  \end{minipage}
  \begin{minipage}[b]{0.195\linewidth}
    \centering Ours
  \end{minipage}
  \begin{minipage}[b]{0.594\linewidth}
    \centering
    \centerline{\includegraphics[height=3.41cm]{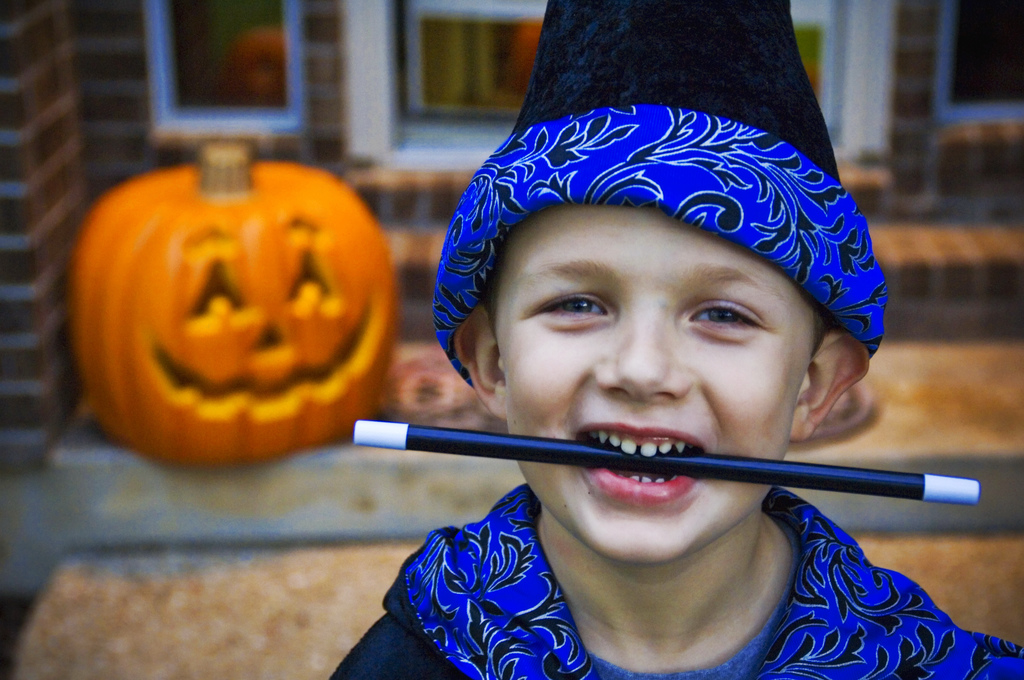}}
  \end{minipage}
  \begin{minipage}[b]{0.195\linewidth}
    \centering
    \centerline{\includegraphics[height=3.41cm]{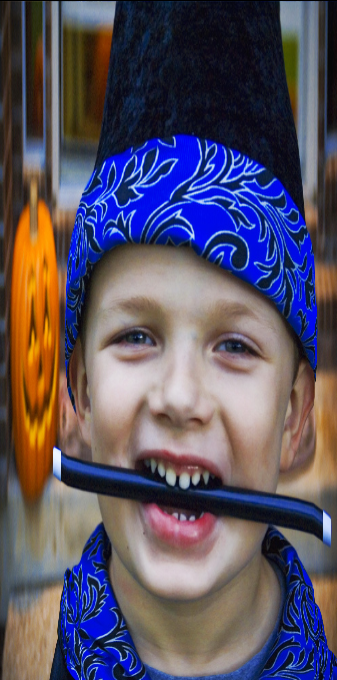}}
  \end{minipage}
  \begin{minipage}[b]{0.195\linewidth}
    \centering
    \centerline{\includegraphics[height=3.41cm]{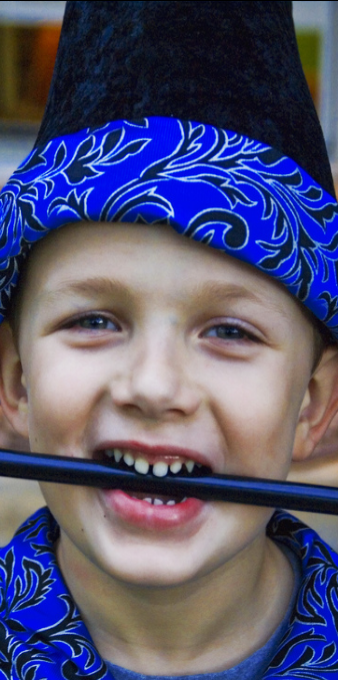}}
  \end{minipage}
  \begin{minipage}[b]{0.594\linewidth}
    \centering
    \centerline{\includegraphics[height=3.41cm]{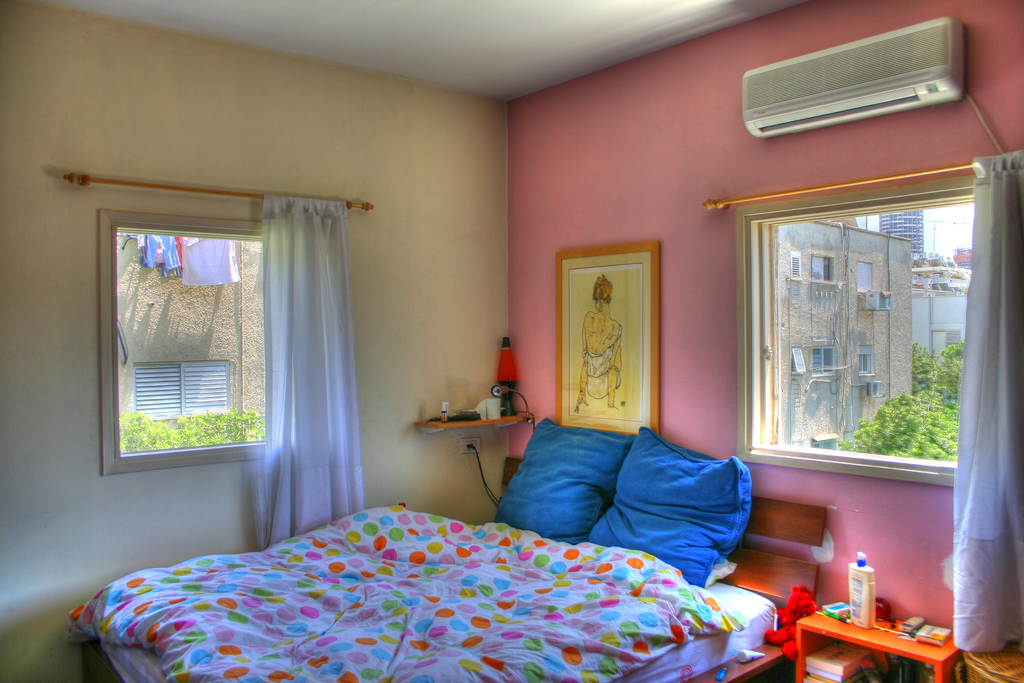}}
  \end{minipage}
  \begin{minipage}[b]{0.195\linewidth}
    \centering
    \centerline{\includegraphics[height=3.41cm]{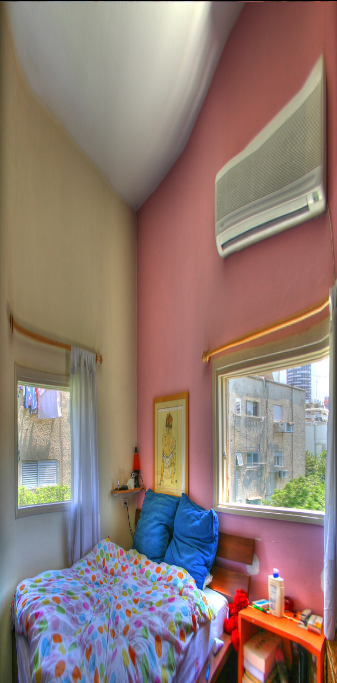}}
  \end{minipage}
  \begin{minipage}[b]{0.195\linewidth}
    \centering
    \centerline{\includegraphics[height=3.41cm]{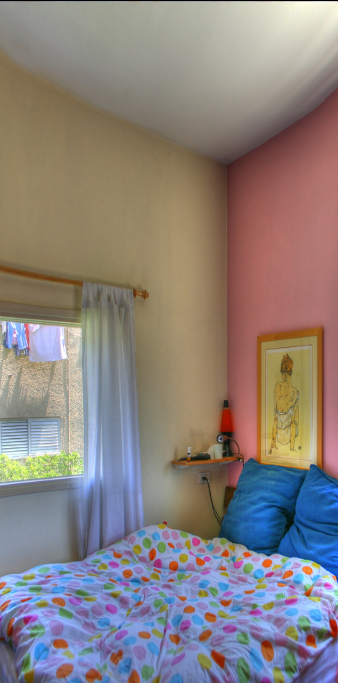}}
  \end{minipage}
  \caption{Source images and their retargeted versions provided by AAD~\cite{AxisAligned:2012} and our method, for a retargeting factor of 0.33.}
  \label{fig:extreme_retargeting}
  \vspace{-0.5cm}
\end{figure}

A measure that quantifies distortion should be sensitive to changes in aspect ratio of large mesh cells, as those will be more noticeable than changes in small cells. This suggests measuring the average change in aspect ratio of cells across the image, weighted by the area and importance of each cell. We define the distortion $D$ by
\begin{equation}
  \label{eqn:somelabel}
  D = \frac{1}{A}\sum_{i=1}^{N} \left[ \frac{\text{max}(h_i', w_i')}{\text{min}(h_i', w_i')}-1 \right] \cdot h_i \cdot w_i \cdot (\Omega_i + \Omega_0),
\end{equation}
where $A$ is the total area of warped image, $N$ is the number of cells, $h_i$ and $w_i$ are the height and width of the $i^{th}$ cell.
We compute the normalised height and width by $h_i'=h_i/h_0$ and $w_i'=w_i/w_0$ where $h_0$ and $w_0$ are the height and width of cells in the uniform mesh.
$\Omega_i$ is the average importance over pixels of the $i^{th}$ cell.
$\Omega_0$ is a tuneable parameter used to penalise the distortion of non-important cells.
The term in square brackets captures the departure of a cell from its original aspect ratio.
Note that the distortion $D$ is not the same as energy $E$ used in the AAD optimisation problem. $E$ is quadratic on the coordinates, and assigns zero weight to unimportant cells, whereas $D$ is linear on the coordinates, and assigns non-zero value to the unimportant cells. We performed experiments using $E$ instead of $D$ as a measure of distortion,
but $D$ was found to be more effective.
Fig.~\ref{fig:distortion} illustrates the effectiveness of $D$ as a measure of distortion: (a) change of $D$ with respect to the warping factor, i.~e. the factor by which the warping changes the source width, whereas (b-d) show sample retargeted images with decreasing distortions on the background buildings.

The intermediate warp is the maximum warp that can be applied to an image while keeping $D<D_t$, where $D_t$ is a tuneable parameter. This is achieved by finding the root of the function ${f(D)=D-D_t}$. Solving this equation is only necessary if the distortion $D$ produced by a warp of the target size is larger than the $D_t$. Otherwise,  warping the image to the target size is considered not to  introduce severe distortions, which is illustrated in the ``Yes" branch of the pipeline in Fig.~\ref{fig:pipeline_full}.

\subsection{Cropping}
\label{ssec:cropping}

If the intermediate warp size is not the same as the target size,
then content-aware cropping is applied (illustrated in the ``No" branch in Fig.~\ref{fig:pipeline_full}).
The regions to be cropped are
distributed on the left and right of the intermediately warped image, aiming to minimise removal of important regions. This is achieved by computing a one-dimensional version of the importance map, and distributing the cropping in such a way as to minimise the removed area under the curve (the one-dimensional map).
For efficiency, the actual cropping is implemented via a final warping mesh, which merges the intermediate warping and cropping into one operation.

Fig.~\ref{fig:distortion} illustrates the trade-off between image distortion and content preservation. Fig.~\ref{fig:distortion} (a) shows the dependence of $D$ on the warping factor.
As the retargeting factor is changed from 1 (unwarped image, with distortion equal zero) to 0.5, the distortion increases monotonically.
Fig.~\ref{fig:distortion} (b), (c), and (d) show retargeted images obtained using different values of the distortion threshold $D_t$, corresponding to the three blue dots in Fig.~\ref{fig:distortion} (a). $D_t$ can be used to tune our system from warping-only to cropping-only, as well as different degrees of hybrid warping-cropping.
Fig.~\ref{fig:distortion} (b) shows warp-only retargeting, which preserves
all of the original content but has severe distortions in the background houses. Fig.~\ref{fig:distortion} (d) shows crop-only retargeting, which produces zero distortions but introduces severe content loss. Fig.~\ref{fig:distortion} (c) shows an image with a good balance between warp and crop, which preserves most of the important content while keeping distortions low.

\begin{table}[tpb]
  \centering
  \begin{tabular}{lr}
    \hline
    Method                              & Normalised Votes \\
    \hline
    SCL                                 &    0.429 \\
    BEL~\cite{beltrami}                 &    0.545 \\
    CycleIR~\cite{CycleIR2019}          &    0.480 \\
    AAD~\cite{AxisAligned:2012}         &    0.475 \\
    Ours                                &   \textbf{0.572} \\
    \hline
  \end{tabular}
  \caption{Normalised votes in our RetargetMe user study. Our method achieves superior performance. Apart from BEL, all methods are fully automatic.}
\label{tab:votes}
\vspace{-0.5cm}
\end{table}

\section{RESULTS}
\label{sec:results}

This section compares our method against recent retargeting work.
We included in our quality analysis only methods in peer reviewed publications and for which we had data available.
Though we do not directly compare against methods in the original RetargetMe study~\cite{RetargetMe:2010} from 2010, we do compare against
AAD, which showed superiority over those approaches~\cite{AxisAligned:2012}. Experimental results show that our method
outperforms all other methods included in our analyses.

\subsection{Implementation Details}
\label{ssec:implementation_details}

Semantic segmentation was performed using a Deeplabv3+ model with a MobileNetV2 backbone \cite{deeplabv3plus2018} pre-trained on the  Pascal VOC 2012 dataset \cite{pascalV2012}.
The semantic segmentation pixel percentage threshold was tuned to 5\%.
For saliency detection, we used a pre-trained model described in \cite{SaliencyContextCodec2020}.
For the distortion measure, we tuned $\Omega_0=1$ and $D_t=1$.

\begin{table}[tpb]
  \centering
  \begin{tabular}{lrrr}
    \hline
    Method     & Time (ms) & Image size & Processor Type \\
    \hline
    CycleIR \cite{CycleIR2019}  & 370 &   1024$\times$768 & GPU \\
    BEL \cite{beltrami}  & 1,000 & 615$\times$461 & CPU \\
    Ref. \cite{selfplay2020}  &  9,900 & 640$\times$480 & CPU \\
    Ref. \cite{resizingByReconstruction2019}  & 120,000 &  640$\times$480 & N/A \\
    Ref. \cite{weaklyAndSelfSupervised2017}  & 1,000 &  224$\times$224 & GPU \\
    Ref. \cite{ahmadi2019contextaware}  & 15,320 & 1024$\times$813 & GPU \\
    Ours  &   \textbf{120} & \textbf{2152$\times$1534} & GPU \\
    \hline
  \end{tabular}
\caption{Reported retargeting execution times on desktop, for different methods. This is for illustrative purposes only, as hardware differences are not accounted for, and a precise timing benchmark is beyond the scope of the present work.}
\label{tab:timings}
\end{table}

\subsection{Visual Comparison}
\label{ssec:visual_comparison}

Fig.~\ref{fig:comparisons} presents results for sample images selected from the RetargetMe dataset \cite{RetargetMe:2010}. On the left is the source image, and each of the other columns shows target images obtained from a variety of methods: SCL corresponds a uniform stretching operation. BEL is a non-fully automatic method that uses hand-crafted importance maps \cite{beltrami}. AAD denotes the warp-only method described in~\cite{AxisAligned:2012}
but employing importance maps produced by our method, as described in Section \ref{ssec:importance_map} (instead of using the hand-crafted maps utilised in \cite{AxisAligned:2012}). CycleIR is a recent DNN-based approach \cite{CycleIR2019}.
Among the retargeting approaches compared in Fig.~\ref{fig:comparisons}, our method attains the best balance between keeping important content and reducing distortions. Fig.~\ref{fig:extreme_retargeting} shows examples of images retargeted by a factor of 0.33 using AAD and our proposed method.

\begin{table}[tpb]
  \centering
  \begin{tabular}{lrrr}
    \hline
    Platform          & Importance Map  & Warp \& Crop    & Total\\
    \hline
    Desktop           & 85              & 35         & 120 \\
    Mobile Device     & 3000            & 130        & 3130  \\
    \hline
  \end{tabular}
  \caption{Retargeting execution times (ms) for an image of size 2152$\times$1534 using our method. All times are measured on CPU, except from that of Importance Map on desktop, which uses a GPU.}
\label{tab:timingsHybrid}
\end{table}

\subsection{User Study}
\label{ssec:user_study}

To compare the quality of our method with those of the selected approaches, we performed a user study using Amazon Mechanical Turk.
Our study compared images retargeted to 0.5 of their original width only.
294 surveys were used, after rejecting 54 that did not pass a hidden sentinel test. Each retargeting method was compared 3564 times on average.
Given that there are small variations in the number of times each method was sampled (due to the finite number of randomly selected pair-wise comparisons),
we performed vote normalisation by dividing the number of votes that each method received by the number of times it was accessed.
Table~\ref{tab:votes} shows the results for the five selected methods, with ours gaining the largest number of votes.

\subsection{Execution Times}
\label{ssec:execution_times}

Though a precise benchmark of execution times is beyond the scope of this work, it is illustrative to compare reported execution times in recent retargeting literature. Table~\ref{tab:timings} shows one such comparison (it includes more entries than the user study of Table~\ref{tab:votes} as timing data was more readily available than the retargeted images needed for our user study.)
For the desktop experiments, we used an NVIDIA GeForce RTX 2080 GPU, and an Intel i9-9820X CPU. For mobile device, we used a Samsung Galaxy Fold (SM-F900).
Note that importance maps need to be generated only once per image, regardless of the target size, and can therefore be done as a pre-processing step. This allows very fast retargeting (35 ms on desktop, and 130 ms on mobile device) to any size once the importance map has been computed.

\section{CONCLUSION}
\label{sec:conclusions}

We have presented a fully automated retargeting method. It combines the
capabilities of DNNs for importance map generation, the efficiency of a warping algorithm~\cite{AxisAligned:2012}, and a novel approach to quantify and limit warping distortions.
By varying a single parameter, namely a distortion threshold, our system can be tuned to perform retargeting by warping-only, cropping-only, or a continuous range of hybrid scenarios in between.
When fixing the distortion threshold to a fine-tuned value for hybrid warp-cropping, our system produces state-of-the-art quality, as demonstrated in a user study on the RetargetMe benchmark. Additionally, our system runs in a fraction of the time of recent retargeting approaches, and has been tested on mobile devices where, after a pre-processing step, retargets images in real-time.
In future work, we plan to optimise for speed, and to improve the accuracy of, the importance map generation models.

\newpage

\clearpage

\bibliographystyle{IEEEbib}
\bibliography{refs}

\end{document}